\documentclass[journal]{IEEEtran}
\usepackage[dvips]{graphicx}
\usepackage{epsfig,multirow}
\usepackage{amsmath,amssymb,bm}
\usepackage{amsfonts,dsfont,color,bbm}
\usepackage{caption}
\usepackage{subcaption}
\usepackage{cite}

\renewcommand{\vec}[1]{\mbox{\boldmath${#1}$}}
\newcommand{\norm}[1]{\left|\left|#1\right|\right|}

\newcommand{\pf}[1]{\mbox{$f(#1,x_{t-a}^{t-1})$}}
\newcommand{\fw}[1]{\mbox{$\un{f}_{w}(#1)$}}
\newcommand{\ph}[1]{\mbox{$h(#1,x_{t-a}^{t-1})$}}
\newcommand{\fx}{\mbox{$\un{f}_{x}(x_{t-a}^{t-1})$}}
\newcommand{\sfxn}{\mbox{$f_{n}(x_{t-a}^{t-1})$}}
\newcommand{\sfx}{\mbox{$f_{x}(x_{t-a}^{t-1})$}}
\newcommand{\un}{\vec}
\newcommand{\uw}{\un{w}}
\newcommand{\ua}{\un{\beta}}
\newcommand{\uth}{\mbox{$\un{\theta}$}}
\newcommand{\xp}[1]{\mbox{$\hat{x}_{#1}[t]$}}
\newcommand{\be}{\begin{equation}}
\newcommand{\ee}{\end{equation}}
\newcommand{\bea}{\begin{eqnarray}}
\newcommand{\eea}{\end{eqnarray}}
\newcommand{\nn}{\nonumber}
\newcommand{\ei}{\end{itemize}}
\newcommand{\bi}{\begin{itemize}}

\newcommand{\Real}{{\mathbbm{R}}}

\DeclareMathOperator*{\argmin}{arg\,min}

\AtBeginDocument{
\addtolength{\abovedisplayskip}{-0.8ex}
\addtolength{\abovedisplayshortskip}{-0.8ex}
\addtolength{\belowdisplayskip}{-0.8ex}
\addtolength{\belowdisplayshortskip}{-0.8ex}
\addtolength{\belowcaptionskip}{-3.6ex}
}

\begin{document}

\title{A Unified Approach to Universal Prediction: Generalized Upper and Lower Bounds}

\author{N. Denizcan Vanli and Suleyman S. Kozat, \textit{Senior Member, IEEE}
\thanks{This work is supported in part by IBM Faculty Award and TUBITAK, Contract no: 112E161.}
\thanks{The authors are with the Department of Electrical and Electronics Engineering, Bilkent University, Bilkent, Ankara 06800, Turkey (e-mail: vanli@ee.bilkent.edu.tr, kozat@ee.bilkent.edu.tr).} }

\maketitle

\begin{abstract}
We study sequential prediction of real-valued, arbitrary and unknown
sequences under the squared error loss as well as the best parametric
predictor out of a large, continuous class of predictors. Inspired by
recent results from computational learning theory, we refrain from any
statistical assumptions and define the performance with respect to the
class of general parametric predictors. In particular, we present
generic lower and upper bounds on this relative performance by
transforming the prediction task into a parameter learning problem. We
first introduce the lower bounds on this relative performance in the
mixture of experts framework, where we show that for any sequential
algorithm, there always exists a sequence for which the performance of
the sequential algorithm is lower bounded by zero. We then introduce a
sequential learning algorithm to predict such arbitrary and unknown
sequences, and calculate upper bounds on its total squared prediction
error for every bounded sequence. We further show that in some
scenarios we achieve matching lower and upper bounds demonstrating
that our algorithms are optimal in a strong minimax sense such that
their performances cannot be improved further. As an interesting
result we also prove that for the worst case scenario, the performance
of randomized algorithms can be achieved by sequential algorithms so
that randomized algorithms does not improve the performance.
\end{abstract}

\vspace{-0.5cm}
\section{Introduction}\label{sec:intro}
In this brief paper, we investigate the generic sequential (online)
prediction problem from an individual sequence perspective using tools
of computational learning theory, where we refrain from any
statistical assumptions either in modeling or on signals
\cite{TNN1,Linder1,Linder2,Warmuth2}. In this approach we have an
arbitrary, deterministic, bounded and unknown signal $\{x[t]\}_{t \geq
  1}$, where $|x[t]| < A < \infty$, and $x[t] \in \Real$. Since we do
not impose any statistical assumptions on the underlying data, we,
motivated by recent results from  sequential learning
\cite{TNN1,Linder1,Linder2,Warmuth2}, define the performance of a
sequential algorithm with respect to a comparison class, where the predictors of the comparison class are formed by observing the the entire sequence in hindsight, under the
squared error loss, i.e.,
\[
\sum_{t=1}^n (x[t]-\hat{x}_s[t])^2 - \inf_{c \in {\cal C}} \sum_{t=1}^n \left(x[t]-\hat{x}_c[t]\right)^2,
\]
for an arbitrary length of data $n$, and for any possible sequence $\{x[t]\}_{t \geq 1}$, where $\hat{x}_s[t]$ is the prediction at time $t$ of any sequential algorithm that has access data from $x[1]$ up to $x[t-1]$ for prediction, and $\hat{x}_c[t]$ is the prediction at time $t$ of the predictor $c$ such that $c \in \cal C$, where ${\cal C}$ represents the class of predictors we ``compete'' against. We emphasize that since the predictors $\hat{x}_c[t]$, $c \in {\cal C}$ have the access to the entire sequence before the processing starts, the minimum squared prediction error that can be achieved with a sequential predictor $\hat{x}_s[t]$ is equal to the squared prediction error of the optimal batch predictor $\hat{x}_c[t]$, $c \in \cal C$. Here, we call the difference in the squared prediction error of the sequential algorithm $\hat{x}_s[t]$ and the optimal batch predictor $\hat{x}_c[t]$, $c \in \cal C$ as the ``regret'' of not using the optimal predictor (or equivalently, not knowing the future). Therefore, we seek for sequential algorithms $\hat{x}_s[t]$ that minimize this ``regret'' or loss for any possible $\{x[t]\}_{t \geq 1}$. We emphasize that this regret definition is for the accumulated sequential cost, instead of the batch cost.

Instead of fixing a comparison class of predictors, we parameterize the comparison classes such that the parameter set and functional form of these classes can be chosen as desired. In this sense, in this paper, we consider the most general class of parametric predictors as our class of predictors $\cal C$ such that the ``regret'' for an arbitrary length of data $n$ is given by
\be \label{eq:regret_det}
\sum_{t=1}^{n} (x[t]-\xp{s})^{2}-\inf_{\un{w} \in \Real^{m}} \sum_{t=1}^{n} \left(x[t]-\pf{\un{w}}\right)^{2},
\ee
where $\pf{\un{w}}$ is a parametric function whose parameters $\un{w}=[w_{1}, \dots, w_{m}]^T$ can be set prior to prediction, and this function uses the data $x_{t-a}^{t-1}$, $t-a \geq 1$ for prediction for some arbitrary integer $a$, which can be viewed as the tap size of the predictor.\footnote{All vectors are column vectors and denoted by boldface lower case letters. For a vector $\un{u}$, $\un{u}^T$ is the ordinary transpose. We denote $x_a^b \triangleq \{x[t]\}_{t=a}^b$.} Although the parameters of the parametric prediction function $\pf{\uw}$ can be set arbitrarily, even by observing all the data $\{x[t]\}_{t \geq 1}$ a priori, the function is naturally restricted to use only the sequential data $x_{1}^{t-1}$ in prediction \cite{Universal,Singer,InfoPaper}.

Since we have no statistical assumptions on the underlying data, the corresponding lower and upper bounds on the regret in \eqref{eq:regret_det} in this sense provide the ``ultimate'' measure of the learning performance for any sequential predictor. We emphasize that lower bounds not only provide the worst-case performance of an algorithm, but also quantify the prediction power of the parametric class. As such, a positive lower bound guarantees the existence of a data sequence having an arbitrary length such that no matter how ``smart'' the learning algorithm is, the performance of this smart algorithm on this sequence will be worse than the class of parametric predictors by at least an order of the ``lower bound''.  Hence if an algorithm is found such that the upper bound of the regret of that algorithm matches with the lower bound, then that algorithm is optimal in a strong minimax sense such that the actual convergence performance cannot be further improved \cite{InfoPaper}. To this end, the minimax sense optimality of different parametric learning algorithms such as the well-known prediction algorithms, least mean squares (LMS) \cite{sayedbook}, recursive least squares (RLS) \cite{sayedbook}, and online sequential extreme learning machine (OS-ELM) of \cite{TNN1} can be determined using the lower bounds provided in this paper. In this sense, the ``rates'' of the corresponding upper and lower bounds are analogous to the VC dimension \cite{TNN2} of classifiers and can be used to quantify the learning performance \cite{Warmuth,TNN1,Linder1,Linder2}.

The mixture of experts framework is previously used in order to derive such upper and lower bounds for the performance of an algorithm. As an example, linear prediction \cite{Universal,InfoPaper,Vovk}, nonlinear models based on locally linear approximations \cite{Singer}, and the learning of an individual noise-corrupted deterministic sequence \cite{Binary} is studied. These results are then extended to the filtering problems \cite{Filtering1,Filtering2}. In this paper on the other hand, we consider a holistic approach and provide upper and lower bounds for the general framework, which was previously missing in the literature.

Our main contribution in this paper is to obtain the generalized lower bounds for a variety of prediction frameworks by transforming the prediction problem to a well-known and studied statistical parameter learning problem \cite{Warmuth2,TNN1,Universal,Singer,InfoPaper}. By doing so, we prove that for any sequential algorithm there always exist some data sequence over any length such that the regret of the sequential algorithm is lower bounded by zero. We further derive lower bounds for important classes of predictors heavily investigated in machine learning literature including univariate polynomial, multivariate polynomial, and linear predictors \cite{Warmuth,Warmuth2,Universal,Singer,InfoPaper,Vovk,Binary,Polynomial}. We also provide a universal sequential prediction algorithm and calculate upper bounds on the regret of this algorithm, and show that we obtain matching lower and upper bounds in some scenarios. As an interesting result we also show that given the regret in \eqref{eq:regret_det} as the performance measure, there is no additional gain achieved by using randomized algorithms in the worst-case scenario.

In Section \ref{sec:lower}, we first present general lower bounds, and then analyze couple of specific scenarios. We then introduce a universal prediction algorithm and calculate the upper bounds on its regret in Section \ref{sec:upper}. In Section \ref{sec:randomized}, we show that in the worst-case scenario, the performance of randomized algorithms can be achieved by sequential algorithms. We finalize our paper by pointing out several concluding remarks.

\vspace{-0.2cm}
\section{Lower Bounds} \label{sec:lower}
In this section, we investigate the worst case performance of sequential algorithms to obtain guaranteed lower bounds on the regret. Hence for any arbitrary length of data $n$, $\{x[t]\}_{t \geq 1}$, we are trying to find a lower bound on the following
\be \label{eq:regret}
\sup_{x_{1}^{n}} \left\{ \sum_{t=1}^{n} (x[t]-\xp{s})^{2}-\inf_{\un{w} \in \Real^{m}} \sum_{t=1}^{n}
(x[t]-\pf{\un{w}})^{2} \right\}.
\ee
For this regret, we have the following theorem which relates the performance of any sequential algorithm to the general class of parametric predictors. While proving this theorem we also provide a generic procedure to find lower bounds on the regret in \eqref{eq:regret} and later use this method to derive lower bounds for parametric classes including the classes of univariate polynomial, multivariate polynomial, and linear predictors \cite{Warmuth,Warmuth2,Universal,Singer,InfoPaper,Vovk,Binary,Polynomial}.

{\bf Theorem 1: }{\em There is no ``best'' sequential algorithm for all sequences for any class in the parametric form $\pf{\un{w}}$, where $\un{w} \in \Real^m$. Given a parametric class there exist always a sequence such that the regret in \eqref{eq:regret} is always lower bounded by some nonnegative value.}

This theorem implies that no matter how smart a sequential algorithm is or how naive the competition class is, it is not possible to outperform the competition class for all sequences. As an example, this result demonstrates that even competing against the class of constant predictors, i.e., the most naive competition class, where $\hat{x}_c[t]$ always predicts a constant value, any sequential algorithm, no matter how smart, cannot outperform this class of constant predictors for all sequences. We emphasize that in this sense, the lower bounds provide the prediction and modeling power of the parametric class.

{\em Proof of Theorem 1: } We begin our proof by pointing out that
finding the ``best'' sequential predictor for an arbitrary and unknown
sequence of $x_1^n$ is not straightforward. Yet, for a specific
distribution on $x_1^n$, the best predictor is the conditional mean on
$x_1^n$ under the squared error \cite{Probability}. Therefore, by this
clever transformation, we are able to calculate the regret in
\eqref{eq:regret} in the expectation sense and prove this theorem.

Since the supremum in \eqref{eq:regret} is taken over all $x_{1}^{n}$, for any distribution $x_{1}^{n}$, the regret is lower bounded by
\bea
\lefteqn{\sup_{x_{1}^{n}} \left(\sum_{t=1}^{n}(x[t]-\xp{s})^{2}-\inf_{\un{w}
\in \Real^{m}} \sum_{t=1}^{n}(x[t]-\pf{\un{w}})^{2} \right)} \nn\\
&& \hspace{-0.7cm} \geq
\underbrace{E_{x_{1}^{n}} \left[ \sum_{t=1}^{n} (x[t]-\xp{s})^{2}-\inf_{\un{w} \in \Real^{m}}
\sum_{t=1}^{n}(x[t]-\pf{\un{w}})^{2} \right]}_{\triangleq L(n)}, \nn
\eea
where expectation is taken with respect to this particular distribution. Hence it is enough to lower bound $L(n)$ to get a final lower bound. By the linearity of the expectation
\begin{align}\label{eq:loss}
  L(n) &= E_{x_{1}^{n}} \left[ \sum_{t=1}^{n} (x[t]-\xp{s})^{2} \right] \nn\\
       & \hspace{0.5cm} - E_{x_{1}^{n}} \left[ \inf_{\un{w} \in \Real^{m}} \sum_{t=1}^{n} (x[t]-\pf{\un{w}})^{2} \right].
\end{align}
The squared-error loss $E\big{[}(x[t]-\xp{s})^{2}\big{]}$ is minimized with the well-known minimum mean squared error (MMSE) predictor given by \cite{Probability}
\be \label{eq:mmse}
 \xp{s} = E \left[ x[t]\big{|}x[t-1],\ldots,x[1] \right]  = E \left[ x[t]\big{|}x^{t-1}_1 \right],
\ee
where we drop the explicit $x_1^n$-dependence of the expectation to simplify the presentation.

Suppose we select a parametric distribution for $x_{1}^{n}$ with parameter vector
$\uth=[\theta_{1}, \ldots, \theta_{m}]$. Then for
\[
E_{x_1^n} \left[ \inf_{\un{w} \in \Real^{m}} \sum_{t=1}^{n} (x[t]-\pf{\un{w}})^{2} \right]
\]
in \eqref{eq:loss}, we use the following inequality
\bea \label{eq:ine}
\lefteqn{E_{\uth}\left[ E_{x_{1}^{n}\big{|}\uth} \left[ \inf_{\un{w} \in \Real^{m}} \sum_{t=1}^{n} (x[t]-\pf{\un{w}})^{2} \right]
\right]} \nn\\
&& \leq E_{\uth}\left[ \inf_{\un{w} \in \Real^{m}}E_{x_{1}^{n}\big{|}\uth} \left[ \sum_{t=1}^{n} (x[t]-\pf{\un{w}})^{2} \right] \right].
\eea
By using \eqref{eq:mmse}-\eqref{eq:ine}, and expanding the expectation we can lower bound $L(n)$ as
\begin{align}\label{eq:above}
  L(n) & \geq E_{\uth}\left[ E_{x_{1}^{n}\big{|}\uth} \left[ \sum_{t=1}^{n} \left(x[t]-E\left[x[t]\big{|}x^{t-1}_1\right]\right)^2 \right] \right] \nn\\
       & \hspace{0.5cm} - E_{\uth} \left[ \inf_{\un{w} \in \Real^{m}}E_{x_{1}^{n}\big{|}\uth} \left[ \sum_{t=1}^{n} (x[t]-\pf{\un{w}})^{2} \right] \right].
\end{align}
The inequality in \eqref{eq:above} is true for any distribution on $x_{1}^{n}$. Hence for a distribution on $x_{1}^{n}$ such that
\be \label{eq:yeni}
 E \left[ x[t] \big{|} x^{t-1}_1, \uth \right]= \ph{\uth},
\ee
with some function $h$, if we can find a vector function $\un{g}(\uth)$ satisfying $\pf{\un{g}(\uth)}=\ph{\uth}$ then the last term in \eqref{eq:above} yields
\bea
\lefteqn{ E_{\uth} \left[ \inf_{\un{w} \in \Real^{m}}E_{x_{1}^{n}\big{|}\uth} \left[ \sum_{t=1}^{n} (x[t]-\pf{\un{w}})^{2} \right] \right] } \nn\\
&& = E_{\uth}\left[ E_{x_{1}^{n} \big{|} \uth} \left[ \sum_{t=1}^{n} (x[t]-\ph{\uth})^{2} \right] \right]. \nn
\eea
Thus \eqref{eq:above} can be written as
\bea
\lefteqn{L(n) \geq E_{\uth}\left[ E_{x_{1}^{n}\big{|}\uth} \left[ \sum_{t=1}^{n} \left(x[t]-E\left[x[t]\big{|}x^{t-1}_1\right]\right)^{2} \right] \right]} \nn\\
&& \hspace{0.5cm} - E_{\uth}\left[ E_{x_{1}^{n}\big{|}\uth} \left[ \sum_{t=1}^{n} \left(x[t]-E\left[x[t]\big{|}x^{t-1}_1,\uth \right] \right)^{2} \right] \right], \nn
\eea
which is by definition of the MMSE estimator is always lower bounded by zero, i.e.,
\[
 L(n) \geq 0.
\]
By this inequality we conclude that for predictors of the form
$\pf{\uw}$ for which this special parametric distribution, i.e., $\uw
= \un{g}(\uth)$ exists, the best sequential predictor will be always
outperformed by some predictor in this class for some sequence
$x_{1}^{n}$. Hence there is no ``best'' algorithm for all sequences
for any class in this parametric form. The question arises if a
suitable distribution on $x_{1}^{n}$ can be found for a given
$\pf{\uw}$ such that $\pf{\un{g}(\uth)}=\ph{\uth}$ with a suitable
transformation $\un{g}(\uth)$.

Suppose $\pf{\uw}$ is bounded by some $M \in R^+$ with $M < \infty$ for all $|x[t]| \leq A$, i.e., $|\pf{\uw}| \leq M$. Then, given $\theta$ from a beta distribution with parameters $(C,C)$,   $C \in R^+$, we generate
a sequence $x_{1}^{n}$ such that $x[t]=\frac{A}{M} \pf{\uw}$  with probability $\theta$
and $x[t]=-\frac{A}{M} \pf{\uw}$ with probability $(1-\theta)$. Then
\[
E \left[ x[t]\big{|}x^{t-1}_1,\theta \right]=\frac{A}{M}(2\theta-1)\pf{\uw}.
\]
Hence, this concludes the proof of the Theorem 1. \hfill $\square$

As an important special case, if we use the restricted functional form $\pf{\uw}$ so that $\pf{\uw}$ is separable, then the prediction problem is transformed to a parameter estimation problem. The separable form is given by
\[
\pf{\uw}=\fw{\uw}^{T}\fx,
\]
where $\fw{\uw}$ and $\fx$ are vector functions of size $m \times 1$ for some integer $m$. Then \eqref{eq:yeni} can be written as
\[
E \left[ x[t]\big{|}x^{t-1}_1,\un{\theta} \right] = \fw{\un{g}(\uth)}^{T}\fx,
\]
where $\fw{\un{g}(\uth)} = \frac{A}{M}(2\theta-1) \fw{\uw}$. Denoting $\un{f}_n(\uw) \triangleq \frac{A}{M} \fw{\uw}$ as the normalized prediction function, and after some algebra \eqref{eq:above} is obtained as
\begin{align}
  & L(n) \geq \nn\\
  & E_{\uth} \hspace{-0.15cm} \left[ E_{x_{1}^{n}\big{|}\uth} \hspace{-0.15cm} \left[ \hspace{-0.05cm} \sum_{t=1}^{n} \hspace{-0.05cm} \left( x[t] \hspace{-0.05cm} - \hspace{-0.05cm} E \hspace{-0.05cm} \left[(2\theta \hspace{-0.05cm} - \hspace{-0.05cm} 1)\big{|}x^{t-1}_1\right]^{T} \hspace{-0.15cm} \un{f}_n(\uw)^{T} \hspace{-0.1cm} \fx\right)^{ \hspace{-0.1cm} 2} \hspace{-0.08cm} \right] \hspace{-0.02cm} \right] \nn\\
  & \hspace{0.15cm} - E_{\uth}\left[ E_{x_{1}^{n}\big{|}\uth} \left[ \sum_{t=1}^{n} \left( x[t]-(2\theta-1)\un{f}_n(\uw)^{T}\fx \right)^{2} \right] \right], \nn
\end{align}
so that the regret of the sequential algorithm over the best prediction function is due to the regret attained by the sequential algorithm while learning the parameters of the prediction function, i.e, the parameters of the underlying distribution. To illustrate this procedure, we investigate the regret given in \eqref{eq:regret} for three candidate function classes that are widely studied in computational learning theory.

\vspace{-0.4cm}
\subsection{$m$th-order Univariate Polynomial Prediction}
For a $m$th order polynomial in $x[t-1]$ the regret is given by
\be \label{eq:regretpoly}
\sup_{x_{1}^{n}} \hspace{-0.05cm} \left\{ \hspace{-0.05cm} \sum_{t=1}^{n} (x[t]-\xp{s})^{2} \hspace{-0.05cm} - \hspace{-0.2cm} \inf_{\un{w} \in \Real^{m}}
\sum_{t=1}^{n} \hspace{-0.05cm} \left( \hspace{-0.05cm} x[t]- \hspace{-0.1cm} \sum_{i=1}^{p}w_{i}x^{i}[t-1] \right)^{ \hspace{-0.15cm} 2} \hspace{-0.1cm} \right\},
\ee
where $\xp{s}$ is the prediction at time $t$ of any sequential algorithm that has access data from $x[1]$ up to $x[t-1]$ for prediction, $\uw=[w_{1}, \ldots, w_{m}]^{T}$ is the parameter vector, $x^{i}[t-1]$ is the $i$th power of $x[t-1]$.

Since $\sum_{i=1}^{m}w_{i}x^{i}[t-1]=w_{1}x[t-1]$ with appropriate selection of $\uw$, considering the following distribution on $x_1^n$, we can lower bound the regret in \eqref{eq:regretpoly}. Given $\theta$ from a beta distribution with parameters $(C,C)$, $C \in R^+$, we generate a sequence $x_{1}^{n}$ having only two values, $A$ and $-A$ such that $x[t]=x[t-1]$ with probability $\theta$ and $x[t]=-x[t-1]$ with probability $(1-\theta)$. Then
\[
E \left[ x[t] \big{|} x^{t-1}_1, \theta \right] = (2\theta-1) x[t-1],
\]
giving $\ph{\uth}=(2\theta-1)x[t-1]$. Since the MMSE given $\theta$ is linear in $x[t-1]$, the optimum $\uw$ that minimizes the accumulated error for this distribution is $\uw=[(2\theta-1),0, \ldots,0]^{T}$. After following the lines in \cite{Universal}, we obtain a lower bound of the form $O(\ln(n))$.

\vspace{-0.4cm}
\subsection{Multivariate Polynomial Prediction}
Suppose the prediction function is given by $\un{w}^{T}\fx=\sum_{k=1}^{m}w_{k}f_{k}(x_{t-r}^{t-1})$, where each $f_{k}(x_{t-r}^{t-1})$ is a multivariate polynomial function (as an example $f_{k}(x_{t-r}^{t-1}) = \frac{x[t-1]x^2[t-2]}{x[t-3]}$), and regret is taken over all $\un{w}=[w_{1},\ldots,w_{m}]^{T} \in \Real^{m}$, i.e.,
\be
\sup_{x_{1}^{n}} \left\{ \sum_{t=1}^{n} (x[t]-\xp{s})^{2}-\inf_{\un{w} \in \Real^m} \sum_{t=1}^{n} \left(x[t]-\un{w}^{T}\fx\right)^{2} \right\}, \nn
\ee
where $\xp{s}$ is the prediction at time $t$ of any sequential algorithm that has access data from $x[1]$ up to $x[t-1]$ for prediction, and $\un{w}$ is the parameter for prediction.

We emphasize that this class of predictors are not only the super set of univariate polynomial predictors, but also widely used in many signal processing applications to model nonlinearity such as Volterra filters \cite{Polynomial}. This filtering technique is attractive when linear filtering techniques do not provide satisfactory results, and includes cross products of the input signals.

Since $\sum_{k=1}^{m}w_{k}f_{k}(x_{t-r}^{t-1}) = w_{1}f_{1}(x_{t-r}^{t-1})$ with an appropriate selection of $\un{w}$ and redefinition of $f_{1}(x_{t-r}^{t-1})$, we define the following parametric distribution on $x_{1}^{n}$ to obtain a lower bound. Given $\theta$ from a beta distribution with parameters $(C,C)$, $C \in R^+$, we generate a sequence $x_{1}^{n}$ having only two values, $A$ and $-A$, such that $x[t]=\sfxn$ with probability $\theta$ and $x[t]=-\sfxn$ with probability $(1-\theta)$, where $\sfxn=\frac{Af_{1}(x_{t-r}^{t-1})}{M}$, i.e. normalized version of $f_{1}(x_{t-r}^{t-1})$. Thus, given $\theta$, $x_{1}^{n}$ forms a two-state Markov chain with transition probability $(1-\theta)$. Then
\[
E\left[ x[t]\big{|}x^{t-1}_1,\theta \right]=(2\theta-1)\sfxn.
\]
The lower bound for the regret is given by
\begin{align}
  L(n) & =  E \left[ (x[t]-(2\hat{\theta}-1)\sfxn)^{2} \right] \nn\\
       & \hspace{0.5cm} - E \left[ (x[t]-(2\theta-1)\sfxn)^{2} \right], \nn
\end{align}
where $\hat{\theta}=E[\theta|x^{t-1}_1]$. After some algebra we achieve
\begin{align}
    L(n) & = - 4 E[\hat{\theta} x[t] \sfxn] + 4 E[\theta x[t] \sfxn] \nn\\
         & \hspace{0.5cm} + E[(2\hat{\theta}-1)^{2}] - E[(2\theta-1)^{2}]. \nn
\end{align}
It can be deduced that
\[
\hat{\theta}=E[\theta|x^{t-1}_1] = \frac{t-2-F_{t-2}+C}{t-2+2C},
\]
where $F_{t-2}$ is the total number of transitions between the two states in a sequence of length $(t-1)$, i.e., $\hat{\theta}$ is ratio of  number of transitions to time period. Hence,
\begin{align}
  & E[\hat{\theta} x[t] \sfxn] = E \left[ \frac{t-2-F_{t-2}+C}{t-2+2C} x[t] \sfxn \right] \nn\\
  & \hspace{0.5cm} = \frac{(t-2+C) E[x[t] \sfxn] - E[F_{t-2} x[t] \sfxn]}{t-2+2C} \nn\\
  & \hspace{0.5cm} = -\frac{1}{t-2+2C}E[(1-\theta)(t-2) x[t] \sfxn] \nn\\
  & \hspace{0.5cm} = \frac{t-2}{t-2+2C}E[\theta x[t] \sfxn], \nn
\end{align}
where the third line follows from
\[
E[x[t] \sfxn]=E[(2\theta-1)A^{2}]=0,
\]
and $E[F_{t-2}|x[t] \sfxn]=(t-2)(1-\theta)$ since $F_{t-2}$ is a binomial random
variable with parameters $(1-\theta)$ and size $(t-2)$. Thus we obtain
\begin{align}
  L(n) & \hspace{-0.05cm} = \hspace{-0.05cm} -4\frac{t-2}{t-2+2C}E[\theta x(t)\sfxn] \hspace{-0.05cm} + \hspace{-0.05cm} 4 E[\theta x(t)\sfxn] \nn\\
       & \hspace{0.5cm} + E[(2\hat{\theta}-1)^{2}] - E[(2\theta-1)^{2}]. \nn
\end{align}
After this line the derivation follows similar lines to \cite{InfoPaper}, giving a lower bound of the form $O(\ln(n))$ for the regret.

\vspace{-0.4cm}
\subsection{$k$-ahead $m$th-order Linear Prediction}
The regret in \eqref{eq:regret} for $k$-ahead $m$th-order linear prediction is given by
\be \label{eq:regretp}
\sup_{x_{1}^{n}} \left\{ \sum_{t=1}^{n} (x[t] - \xp{s})^{2} - \hspace{-0.2cm} \inf_{\un{w} \in \Real^{m}} \sum_{t=1}^{n} \left( x[t] - \uw^{T}\un{x}[t-k] \right)^{2} \right\},
\ee
where $\hat{x}_s[t]$ is the prediction at time $t$ of any sequential
algorithm that has access data from $x[1]$ up to $x[t-k]$ for prediction for some integer $k$, $\uw=[w_{1}, \ldots, w_{m}]^{T}$ is the parameter vector, and $\un{x}[t-k]=[x[t-k], \dots,x[t-k-m+1]]^{T}$.

We first find a lower bound for $k$-ahead first-order prediction where $\uw^{T}\un{x}[t-k]=wx[t-k]$. For this purpose we define the following parametric distribution on $x_{1}^{n}$ as in \cite{Universal}. Given $\theta$ from a beta distribution with parameters $(C,C)$,  $C \in R^+$, we generate a sequence $x_{1}^{n}$ having only two values, $A$ and $-A$, such that $x[t]=x[t-k]$ with probability $\theta$ and $x[t]=-x[t-k]$ with probability $(1-\theta)$. Thus, given $\theta$, $x_{1}^{n}$ forms a two-state Markov chain with transition probability $(1-\theta)$. Then,
\[
E\left[ x[t] \big{|} x^{t-k}_1, \theta \right] = (2\theta-1) x[t-k]
\]
giving $\ph{\uth}=(2\theta-1)x[t-k]$ and $\un{g}(\uth)=(2\theta-1)$. After this point the derivation exactly follows the lines in \cite{Universal} resulting a lower bound of the form $O(\ln(n))$.

For $k$-ahead $m$th-order prediction, we generalize the lower bound
obtained for $k$-ahead first-order prediction and following the lines
in \cite{Universal}, we obtain a lower bound of the form $O(m\ln(n))$.

We next derive upper bounds for a universal sequential prediction algorithm.

\vspace{-.4cm}
\section{A Comprehensive Approach to Regret Minimization} \label{sec:upper}
In this section, we introduce a method which can be used to predict a bounded, arbitrary, and unknown sequence. We derive the upper bounds of this algorithm such that for any sequence $x_1^n$, our algorithm will not perform worse than the presented upper bounds. In some cases, by achieving matching upper and lower bounds, we prove that this algorithm is optimal in a strong minimax sense such that the worst-case performance cannot be further improved.

We restrict the prediction functions to be separable, i.e., $\pf{\uw}=\fw{\uw}^{T}\fx$, where $\fw{\uw}$ and $\fx$ are vector functions of size $m \times 1$ for some $m$ integer. To avoid any confusion we simply denote $\ua \triangleq \fw{\uw}$, where $\ua \in \Real^{m}$. Hence, the same prediction function can be written as $\pf{\uw}=\ua^{T}\fx$.

If the parameter vector $\ua$ is selected such that the total squared prediction error is minimized over a batch of data of length $n$, then the coefficients are given by
\[
\ua^*[n] = \argmin_{\ua \in \Real^{m}} \sum_{t=1}^n \left(x[t] - \ua^T \fx\right)^2.
\]
The well-known least-squares solution to this problem is given by $\ua^*[n] = (R_{\un{f}\un{f}}^n)^{-1} r_{x\un{f}}^n$, where
\[
R_{\un{f}\un{f}}^n \triangleq \sum_{t=1}^n \fx \fx^T
\]
is invertible and
\[
r_{x \un{f} }^n \triangleq \sum_{t=1}^n x[t] \fx.
\]
When $R_{\un{f}\un{f}}^n$ is singular, the solution is no longer unique, however a suitable choice can be made using, e.g. pseudoinverses.

We also consider the more general least-squares (ridge regression) problem that arises in many signal processing problems, and whose total squared prediction error is minimized over a batch of data of length $n$ with
\begin{align}
 \ua^*[n] & = \argmin_{\ua \in R^{m}} \left\{ \sum_{t=1}^n \left(x[t] - \ua^T \fx\right)^2 + \delta \norm{\ua}^2 \right\}, \nn\\
          & = \left[ R_{\un{f}\un{f}}^{n} + \delta \un{I} \right]^{-1}r_{x\un{f}}^n. \nn
\end{align}
We define a universal predictor $\tilde{x}_u[n]$, as
\[
\tilde{x}_u[n] = \ua_u[n-1]^T \un{f}(x_{n-a}^{n-1}),
\]
where
\[
\ua_u [n] = \ua^*[n] = \left[ R_{\un{f} \un{f}}^{n} + \delta I \right]^{-1} r_{x\un{f}}^{n},
\]
and $\delta>0$ is a positive constant.

{\bf Theorem 2: }{\em The total squared prediction error of the $m$th-order universal predictor for any bounded arbitrary sequence of $ \{x[t]\}_{t \geq 1}$, $|x[t]| \leq A$, having an arbitrary length of $n$ satisfies
\begin{align}
    \sum_{t=1}^{n}(x[t]-\tilde{x}_u[t])^{2} &  \hspace{-0.05cm} \leq \hspace{-0.15cm} \min_{\ua \in R^{m}} \hspace{-0.15cm} \left\{  \hspace{-0.05cm} \sum_{t=1}^{n} (x[t] \hspace{-0.05cm} - \hspace{-0.05cm} \ua^{T} \hspace{-0.1cm} \fx)^{2} \hspace{-0.05cm} + \hspace{-0.05cm} \delta \norm{\ua}^{2} \hspace{-0.05cm} \right\} \nn\\
    & \hspace{0.5cm} + A^{2} \ln \big{|} I+R_{\un{f}\un{f}}^{n} \delta^{-1} \big{|} . \nn
\end{align} }

Theorem 2 indicates that the total squared prediction error of the $m$th-order universal predictor is within $O(m\ln(n))$ of the best batch $m$th-order parametric predictor for any individual sequence of $\{x[t]\}_{t \geq 1}$. This result implies that in order to learn $m$ parameters, the universal algorithm pays a regret of $O(m\ln(n))$, which can be viewed as the parameter regret. After we prove Theorem 2, we apply Theorem 2 to the competition classes discussed in Section \ref{sec:lower}.

{\em Proof of Theorem 2: } We prove this result for a scalar prediction function such that $\fx=\sfx$ to avoid any confusions. Yet for a vector prediction function of $\fx$, one can follow the exact same steps in this proof with vector extensions of the Gaussian mixture.

The derivations follow similar lines to \cite{Warmuth,Universal}, hence only main points are presented. We first define a function of the loss, namely the ``probability'' for a predictor having parameter $\beta$ as follows
\be
  P_{\beta}(x_{1}^n) = \exp \left(-\frac{1}{2h} \sum_{k=1}^n (x[k]-\beta \sfx)^2 \right), \nn
\ee
which can be viewed as a probability assignment of the predictor with parameter $\beta$ to the data $x[t]$, for $1 \leq t \leq n$, induced by performance of $\beta$ on the sequence $x_1^n$.  We then construct a universal estimate of the probability of the sequence $x_1^n$, as an a-priori weighted mixture among all of the probabilities, i.e., $P_u(x_1^n) = \int_{-\infty}^{\infty} p(\beta) P_{\beta}(x_1^n) d\beta$, where $p(\beta)$ is an a-priori weight assigned to the parameter $\beta$, and is selected as Gaussian in order to obtain a closed form bounds, i.e., $p(\beta) = \frac{1}{\sqrt{2 \pi}\sigma} \exp\left\{-\frac{\beta^2}{2\sigma^2}\right\}$.

Following similar lines to \cite{InfoPaper} with a predictor of $\beta\sfx$ we obtain
\[
P_u(x_n|x^{n-1}) = \gamma \exp\left\{\frac{-1}{2h} \gamma^2 \left( x[n] - \beta[n-1]  f(x_{n-a}^{n-1}) \right)^2 \right\},
\]
where $\gamma \triangleq \sqrt{(R_{ff}^{n-2}+\delta)/(R_{ff}^{n-1}+\delta)}$. If we could find another Gaussian satisfying $\tilde{P}_u(x^n) \geq P_u(x^n)$, then it would complete the proof of the theorem.

After some algebra we find that the universal predictor is given by
\[
\tilde{x}_u[n] = \gamma^2 \beta^*[n-1]f(x_{n-a}^{n-1}) = \frac{r_{xf}^{n-1}}{R_{ff}^{n-1} + \delta}f(x_{n-a}^{n-1}).
\]
Now we can select the smallest value of $h$ over the region $[-A,A]$, $\tilde{P}_u(x_n|x^{n-1})$ is larger than $P_u(x_n|x^{n-1})$, i.e.,
\begin{align}
  & A \leq \frac{\sqrt{2 h \ln(\gamma) (\gamma^2-1) + \gamma^2 \hat{x}_u[n]^2 (1-\gamma^2)}}{(1-\gamma^2)} \nn \\
  & h \geq \frac{A^2(1-\gamma^2) - \gamma^2 \hat{x}_u[n]^2} {-2 \ln (\gamma)}, \nn
\end{align}
which must hold for all values of $\hat{x}_u[n] \in [-A,A]$. Therefore $h \geq A^2 \frac{(1-\gamma^2)}{-2 \ln (\gamma)}$, where $\gamma < 1$.  Note that for $0 < \gamma <1 $ we have $0 < \frac{(1-\gamma^2)}{-2 \ln \gamma} < 1$, which implies that we must have $h \geq A^2$ to ensure that $\tilde{P}_u \geq P_u$. In fact, since this bound on the value of $h$ depends upon the value of  $\gamma$ and $\hat{x}_u[n]$, and is only tight for $\gamma \rightarrow 1$, and $\hat{x}_u[n]=0$, then the restriction that $|x[n]|<A$ can actually be occasionally violated, as long as $\tilde{P}_u \geq P_u$ still holds. \hfill $\square$

To illustrate this procedure, we investigate the upper bound for the regret in \eqref{eq:regret} for the same candidate function classes as we also investigated in Section \ref{sec:lower}.

\vspace{-0.3cm}
\subsection{$m$th-order Univariate Polynomial Predictor }
For a $m$th order polynomial in $x[t-1]$, the prediction function is given by $\pf{\uw}=\ua^{T}\fx=\ua^{T}\un{m}[t-1]$ where $\un{m}[t-1]=[x[t-1],\ldots,x^{m}[t-1]]^{T}$, i.e. the vector of powers of $x[t-1]$. After replacing $R_{\un{f}\un{f}}^n = R_{\un{m}\un{m}}^{n}=\sum_{t=1}^n \un{m}[t-1]\un{m}[t-1]^{T}$ and $r_{x \un{f} }^n =r_{x\un{m}}^n= \sum_{t=1}^n x[t]\un{m}[t-1]$, we obtain an upper bound
\begin{align}
    \sum_{t=1}^{n}(x[t]-\tilde{x}_u[t])^{2} & \hspace{-0.05cm}  \leq \hspace{-0.15cm} \min_{\ua \in \Real^{m}} \hspace{-0.1cm} \left\{ \hspace{-0.05cm} \sum_{t=1}^{n} (x[t] \hspace{-0.05cm} - \hspace{-0.05cm} \ua^{T} \hspace{-0.1cm} \un{m}[t-1])^{2} \hspace{-0.05cm} + \hspace{-0.05cm} \delta \norm{\ua}^{2} \hspace{-0.1cm}  \right\} \nn\\
    & \hspace{0.5cm} + A^{2} \ln \big{|} I+R_{\un{m}\un{m}}^{n} \delta^{-1} \big{|}, \nn
\end{align}
which implies that
\begin{align}
    \sum_{t=1}^{n}(x[t]-\tilde{x}_u[t])^{2} & \hspace{-0.05cm} \leq \hspace{-0.15cm} \min_{\ua \in \Real^{m}} \hspace{-0.05cm} \left\{ \hspace{-0.05cm} \sum_{t=1}^{n} (x[t] \hspace{-0.05cm} - \hspace{-0.05cm} \ua^{T} \hspace{-0.1cm} \un{m}[t-1])^{2} \hspace{-0.05cm} + \hspace{-0.05cm} \delta \norm{\ua}^{2} \hspace{-0.1cm} \right\} \nn\\
    & \hspace{0.5cm} + A^{2} m \ln \left( 1+\frac{A^{2}n}{\delta} \right). \nn
\end{align}

\vspace{-0.3cm}
\subsection{Multivariate Polynomial Prediction}
The upper bound for a multivariate polynomial prediction function $\fx$ exactly follows the upper bound derivation of $m$th order univariate polynomial predictor giving an upper bound
\vspace{-0.1cm}
\begin{align}
    \sum_{t=1}^{n}(x[t]-\tilde{x}_u[t])^{2} & \hspace{-0.05cm} \leq \hspace{-0.15cm} \min_{\ua \in \Real^{m}} \hspace{-0.1cm} \left\{ \hspace{-0.05cm} \sum_{t=1}^{n} (x[t] \hspace{-0.05cm} - \hspace{-0.05cm} \ua^{T} \hspace{-0.1cm} \fx)^{2} \hspace{-0.05cm} + \hspace{-0.05cm} \delta \norm{\ua}^{2} \hspace{-0.1cm} \right\} \nn\\
    & \hspace{0.5cm} + A^{2} m \ln \left( 1+\frac{A^{2}n}{\delta} \right). \nn
\end{align}

\vspace{-0.6cm}
\subsection{$k$-ahead $m$th-order Linear Prediction}
For $k$-ahead $m$th-order prediction, the prediction class is given by $\pf{\uw}=\ua^{T}\fx=\ua^{T}\un{x}[t-k]$ where $\un{x}[t-k]=[x[t-k],\ldots,x[t-k-m+1]]^{T}$ as before. After replacing $R_{\un{f}\un{f}}^n= R_{\un{x}\un{x}}^{n} =\sum_{t=1}^n \un{x}[t-k]\un{x}[t-k]^{T}$ and $r_{x \un{f} }^n =r_{x\un{x}}^n= \sum_{t=1}^n x[t]\un{x}[t-k]$
with suitable limits we obtain an upper bound
\vspace{-0.1cm}
\begin{align}
    \sum_{t=1}^{n}(x[t]-\tilde{x}_u[t])^{2} & \hspace{-0.05cm} \leq \hspace{-0.15cm} \min_{\ua \in \Real^{m}} \hspace{-0.1cm} \left\{ \hspace{-0.05cm} \sum_{t=1}^{n}
    (x[t] \hspace{-0.05cm} - \hspace{-0.05cm} \ua^{T} \hspace{-0.1cm} \un{x}[t-k])^{2} \hspace{-0.05cm} + \hspace{-0.05cm} \delta \norm{\ua}^{2} \hspace{-0.1cm} \right\} \nn\\
    & \hspace{0.5cm} + A^{2} m \ln \left( 1+\frac{A^{2}n}{\delta} \right). \nn
\end{align}

\vspace{-0.4cm}
\section{Randomized Output Predictions} \label{sec:randomized}
\vspace{-0.1cm}
In this section, we investigate the performance of randomized output algorithms for the worst-case scenario with respect to linear predictors with using the same regret measure in \eqref{eq:regret}. We emphasize that the randomized output algorithms are a super set of the deterministic sequential predictors and the derivations here can be readily generalized to include any prediction class. Particularly, we consider randomized output algorithms $f\left(\un{\theta}(x_1^{t-1}),x_1^{t-1}\right)$ such that the randomization parameters $\un{\theta}\in R^m$ can be a function of the whole past. Hence, a randomized sequential algorithm introduce randomization or uncertainty in its output such that the output also depends on a random element. Note that such methods are widely used in applications involving security considerations. As an example, suppose there are $m$ prediction algorithms running in parallel to predict the observation sequence $\{x[t]\}_{t \geq 1}$ sequentially. At each time $t$, the randomized output algorithm selects one of the constituent algorithms randomly such that the algorithm $k$ is selected with probability $p_{k}[t]$. By definition $\sum_{k=1}^m p_{k}[t]=1 $ and $p_{k}[t]$ may be generated as the combination of the past observation samples $x_1^{t-1}$ and a seed independent from the observations.

For such randomized output prediction algorithms we consider the following time-accumulated prediction error over a deterministic sequence $\{x[t]\}_{t \geq 1}$ as the prediction error,
\vspace{-0.1cm}
\be \label{eq:expran}
    P_{\text{rand}}(n) = \sum_{t=1}^{n} E_{\un{\theta}} \left[ \left( x[t]- f\left(\un{\theta}(x_1^{t-1}),x_1^{t-1}\right) \right)^{2} \right].
\ee
This expectation is taken over all the randomization due to independent or dependent seeds. Hence our general regret can be extended to include this performance measure
\be \label{eq:randregret}
    \sup_{x_1^n} \left\{ P_{\text{rand}}(n) - \min_{\un{w}\in \Real^m}\sum_{t=1}^{n} \left( x[t]-\un{w}^{T}\un{x}[t-1]\right)^2 \right\}.
\ee
Expanding \eqref{eq:expran} we obtain
\vspace{-0.1cm}
\begin{align}
    P_{\text{rand}}(n) & = \sum_{t=1}^{n} \bigg{\{} \left( x[t]-E_{\un{\theta}}\left[ f\left(\un{\theta}(x_1^{t-1}),x_1^{t-1}\right) \right] \right)^2 \nn\\
    & \hspace{1.5cm} + \mbox{Var}_{\un{\theta}} \left(f\left(\un{\theta}(x_1^{t-1}),x_1^{t-1}\right) \right) \bigg{\}}, \nn
\end{align}
noting that $x[t]$ is independent of the randomization. Since $E_{\un{\theta}}\left[f\left(\un{\theta}(x_1^{t-1}),x_1^{t-1}\right) \right]$ is a sequential function of $x_{1}^{t-1}$ and $\mbox{Var}_{\un{\theta}} \left(f\left(\un{\theta}(x_1^{t-1}),x_1^{t-1}\right) \right)$ is always nonnegative, the performance of a randomized output algorithm can be reached by a deterministic sequential algorithm.

Since deterministic algorithms are subclass of randomized output algorithms, upper bounds we derived for $k$-ahead $m$th-order prediction in \eqref{eq:regretp} also hold for \eqref{eq:randregret}. Since we also proved that the lower bound for such linear predictions of $m$th order are in the form of $O(m\ln(n))$, the lower and upper bounds are tight and of the form $O(m\ln(n))$.

\vspace{-0.3cm}
\section{Concluding Remarks} \label{sec:conclusion}
\vspace{-0.1cm}
In this paper, we consider the problem of sequential prediction from a mixture of experts perspective. We have introduced comprehensive lower bounds on the sequential learning framework by proving that for any sequential algorithm, there always exists a sequence for which the sequential predictor cannot outperform the class of parametric predictors, whose parameters are set non-casually. The lower bounds for important parametric classes such as univariate polynomial, multivariate polynomial, and linear predictor classes are derived in detail. We then introduced a universal sequential prediction algorithm and investigated the upper bound on the regret of this algorithm. We also derived the upper bounds in detail for the same important classes that we discussed for lower bounds, where we further showed that this algorithm is optimal in a strong minimax sense for some scenarios. Finally, we have proven that for the worst-case scenario, randomized algorithms cannot provide any improvement in the performance compared to the sequential algorithms.

\renewcommand{\baselinestretch}{.94}

\vspace{-0.3cm}
\bibliographystyle{IEEEtran}
\bibliography{bibliography_brief}

\end{document}